# Dynamic benchmarking framework for LLM-based conversational data capture

Pietro Alessandro Aluffi
pietro@sea.dev
sea.dev

Patrick Zietkiewicz
patrickzietkiewicz@gmail.com
sea.dev

Marya Bazzi
marya@sea.dev
sea.dev

Matt Arderne
matt@sea.dev
sea.dev

Vladimirs Murevics
vladimirs@sea.dev
sea.dev

## Abstract

The rapid evolution of large language models (LLMs) has transformed conversational agents, enabling complex human-machine interactions. However, evaluation frameworks often focus on single tasks, failing to capture the dynamic nature of multi-turn dialogues. This paper introduces a dynamic benchmarking framework to assess LLM-based conversational agents through interactions with synthetic users. The framework integrates generative agent simulation to evaluate performance on key dimensions: information extraction, context awareness, and adaptive engagement. By simulating various aspects of user behavior, our work provides a scalable, automated, and flexible benchmarking approach. Experimental evaluation—within a loan application use case—demonstrates the framework's effectiveness under one-shot and few-shot extraction conditions. Results show that adaptive strategies improve data extraction accuracy, especially when handling ambiguous responses. Future work will extend its applicability to broader domains and incorporate additional metrics (e.g., conversational coherence, user engagement). This study contributes a structured, scalable approach to evaluating LLM-based conversational agents, facilitating real-world deployment.

## 1 Introduction

The emergence of large language models (LLMs) has transformed the landscape of conversational artificial intelligence (AI), enabling human-machine interactions that were previously hard to imagine. These models have achieved remarkable capabilities in natural language understanding and generation [1]. However, as these systems transition from research environments to real-world applications, effectively evaluating their performance becomes increasingly critical, particularly in scenarios involving dynamic, multi-turn conversations [3].

Traditional evaluation frameworks for LLMs often rely on single-task assessments, which fail to capture the complexities of extended dialogues and diverse user behaviors. As Laskar et al. [16] highlight, benchmarks frequently omit aspects of conversational interaction, such as coherence and adaptability to varying user responses. These approaches are also not scalable in their pipeline for assessing conversations, limiting their applicability in dynamic, real-world settings. Traditional evaluation frameworks, such as static benchmark-based assessments (e.g., MMLU [13], HumanEval [5]), often focus on narrow, task-specific metrics and fail to adapt to the evolving nature of conversational interactions. To address some of the limitations of traditional evaluation methods, such as their lack of scalability, inability to capture the complexities of extended dialogues, and failure to assess coherence and adaptability in dynamic interactions, we propose a framework for evaluating LLM-based conversational agents through interactions with synthetic users.

Our approach builds on advances in generative agent simulation and incorporates dynamic benchmarking principles [3]. The focus of this paper is on evaluating the coherence and flow of conversations in LLM-based systems, other dimensions of conversational LLM systems are out of scope for this contribution. In particular, our framework aims to assess data extraction consistency, context awareness, and adaptive engagement to improve user experience. To establish a clear foundation for our approach, we define the core elements of our evaluation framework as follows. *Conversational agent* (also referred to as *agent* throughout the paper) refers to a generative AI conversational system designed to engage in multi-turn dialogues, with the objective of extracting structured information in an adaptive manner, based on user interactions. *Data model* is a structured representation of the required data fields that the agent must collect during interactions, serving as a blueprint to assess the completeness and accuracy of the agent's responses. *Synthetic users* are AI-driven personas that simulate real-world user behaviors. Our framework offers a scalable, automated, and flexible mechanism to (1) incorporate diverse aspects of user behaviors and (2) evaluate an agent against a defined user base. To demonstrate its effectiveness, we test our framework using a loan application scenario. Such scenarios are a relevant context for two reasons. First, conversational agents are increasingly used in customer service interactions such as loan applications. Second, these types of interactions require accurate and complete user profiles, suffering from inconsistent and incomplete user responses. Our framework allows us to assess a large number of conversations and diverse scenarios with minimal manual effort, an important capability in an era of increasing conversational LLM-driven systems.

## 2 Related Work

Rapid advancements in LLMs have significantly enhanced the capabilities of conversational agents, enabling them to perform complex tasks across diverse domains. However, robust evaluation of these models remains a challenge [15]. Recent methodologies have explored game-based evaluation frameworks that employ interactive scenarios to assess the diagnostic capabilities of LLMs [21], [4]. Castillo-Bolado et al.[3] introduced a dynamic benchmarking system for conversational agents, highlighting that while LLMs excel in single-task interactions, they often struggle with complex, multi-task scenarios. These findings underscore a critical limitation of current state-of-the-art benchmarks: they often fail to reflect real-world usage of chat agents, particularly the complexities introduced by conversational dynamics. These challenges can be addressed by evaluating LLMs in realistic, multi-task interactions, assessing critical capabilities such as long-term memory, continual learning, and the ability to integrate information under complex conversational conditions. This approach exposes performance gaps that single-task evaluations typically overlook, offering a more comprehensive understanding of LLMs' strengths and limitations [17].

There is a rich body of literature on conversational agents, and LLMs more specifically, evaluation. Task completion and accuracy metrics are fundamental for evaluating whether LLMs achieve specific objectives, particularly in structured tasks like financial analysis or regulatory compliance [2]. Dialogue flow and coherence metrics, in contrast, focus on the logical progression and naturalness of conversations, highlighting the ability of generative models to maintain discourse consistency across domains [18]. Empathy and personalization metrics measure the model's adaptability to user-specific needs, such as tone modulation and response customization, which are critical for applications like customer support or financial advisory services. These metrics enable LLMs to tailor interactions to individual preferences, enhancing user experience and engagement. For example, personalized investment suggestions and human-like empathy in financial advice can positively influence consumer perceptions of authenticity, thereby improving their intention to engage with such services [24]. Evaluating conversational agents from a user experience perspective emphasizes usability, satisfaction, and engagement. Metrics such as the System Usability Scale (SUS) and interaction-derived data—including session duration, message length, and frequency of user-agent exchanges—offer valuable insights into user experience [7] [12]. Engagement metrics are particularly important in long-form conversations, where maintaining user interest is critical. Logical flow and coherence also play a crucial role in keeping the user engaged. Traditional metrics like BLEU [20] correlate poorly with human judgments, leading researchers to explore alternative evaluation methods [8]. Metrics such as turn-level appropriateness, response relevance, and dialogue quality have been widely adopted, relying on linguistic and structural analyses of conversation transcripts for consistent and reproducible assessments without extensive human intervention [18]. Additionally, graph-based and machine learning approaches, which utilize knowledge graphs and word embeddings, have emerged as powerful techniques to measure semantic coherence [23].

A significant challenge in LLM evaluation is the potential trade-off between scalable benchmarks and human-centric assessments. While automated evaluations provide scalability, subjective user evaluations can provide deeper insights and a more nuanced understanding of human-agent interaction. [10] [25]. As part of this study, we focus on automated evaluation metrics due to their scalability. Our contribution aims to propose a scalable framework that can represent and adapt to the complexity and variability of user behavior.

## 3 Methods

This section describes our approach to assess generative conversational agents across key dimensions aimed at improving user experience and engagement, including information extraction accuracy (e.g., exact matches and field level matches) and completeness [2][14] [6]. The evaluation framework is designed to reflect the challenges posed by diverse user behavior types, using LLM-based synthetic users to create conversations where certain aspects of human behavior (e.g., incomplete or inconsistent answers) are simulated. The synthetic users represent a spectrum of interaction styles, from clear and cooperative responses to ambiguous or adversarial behaviors. By incorporating these varied user types, the framework systematically tests the agent's ability to adapt, maintain coherence, and achieve functional objectives in different conversational contexts. This approach aligns with prior studies emphasizing the importance of dynamic benchmarking systems for conversational agents, which highlight how multi-task scenarios expose gaps in performance that static benchmarks often fail to capture [3], [17].

### 3.1 Simulation Framework

Inspired by Conversational Question Answering systems (CoQA) [22], a large dataset of conversational question–answering exchanges, we develop an evaluation framework to assess an LLM's ability to capture data while handling ambiguous information in responses, in the context of a simplified loan application. The framework proposed systematically evaluates a conversational agent in multi-turn dialogues. The use-case we focus on is that of data collection through conversational LLM-based systems. The framework's core functionality consists of three components: (1) data model that need to be collected from users (denoted $D$); (2) a conversational agent; and (3) synthetic user profiles.

The data model in $D$ acts as the blueprint for data collection, specifying specific fields that the conversational agent must populate. For the purpose of this simulation, we selected 20 data fields which are a key to a common loan application (e.g., name, address, phone number, the full list is in Appendix B). Both the agent and the synthetic users are based on the GPT-4o model from OpenAI [19]. The agent adapts dynamically to the state of the conversation. Its prompt includes the data model to be populated ($D$), the extracted data at each iteration, and a set of instructions for conversing with the user. Synthetic user profiles, further described in Section 3.3, introduce variability by simulating different user traits, including tone and response patterns. Furthermore, they are given a user profile, which contains the information needed to answer the agent as well as the ground truth for evaluation. The iterative process begins by initializing the agent. A synthetic user initiates the conversation with an initial message $u_0$. In each iteration, the agent generates a response $a_t$ based on the current data model and interaction history $H$. The user's message and the agent's response are stored in $H$, and the data model $D$ is updated accordingly. The process continues until the model is completely populated, or the maximum number of steps $T$ is reached. Finally, the populated data model $D$ is compared to the ground truth, and the evaluation metrics described in Section 3.5 are computed. This iterative process is repeated 20 times for each user profile (standard and ambiguous) and for each scenario (one-shot extraction, few-shot extraction).

Algorithm 1 in the appendix contains a summary of the pseudocode for our framework.

### 3.2 Conversational Agent

The agent is designed to interact with the user to gather information and populate a data model with accurate and complete data. At every step of the conversation, the agent is aware of the current state of the data collection, which includes the data already collected, any remaining fields that are incomplete, and those that are marked as *unclear*. This allows the agent to stay aware of what information is missing or needs further clarification in order to offer an improved user experience and accurate extraction.

Example Interaction:
- Agent: "What is your monthly income?"
- User: "It's a combination of various sources, but generally, it falls within the mid-five-figure range annually."
- Agent: "Could you please specify a more precise annual income amount and the currency?"

Depending on the experimental settings, if the user provides ambiguous, incomplete, or conflicting information, the agent either politely asks gentle follow-up questions to clarify the user's input and resolve uncertainties, or it avoids follow-up questions and marks fields as "unclear" if it cannot determine the correct information.

### 3.3 Synthetic user profiles

In order to evaluate the robustness and coherence of our conversational agent, one can simulate aspects of human behavior through synthetic users. We introduce two distinct LLM-based synthetic user profiles that simulate real-world interaction patterns. Each synthetic user presents unique challenges:

*Standard User.* Standard users provide clear, concise, and complete responses that align closely with the agent's questions. These users serve as a baseline for evaluating the agent's efficiency in extracting structured information under ideal conditions. Example Interaction:
- Agent: "Could you please provide your annual income and the currency?"
- User: "My annual income is 60,000 USD."
- Agent: "Thank you. Could you please specify the loan amount you are seeking and the currency?"

*Ambiguous User.* Ambiguous users provide responses that are credible yet vague or incomplete, compelling the agent to seek clarification. Their answers are structured in a way that introduces natural uncertainty or leaves out critical details, as outlined in the provided ambiguity guidelines. These users test the agent's ability to probe for missing information and resolve ambiguities effectively. The synthetic user is instructed to resolve ambiguity if directly prompted by the agent. Example Interaction:
- Agent: Could you please tell me the amount you wish to borrow and the currency?"
- User: "Somewhere in the five-figure range, comparable to a small car's cost."
- Agent: "Could you please specify the exact loan amount you are considering and the currency?"
- User: "25,000 USD."

### 3.4 Evaluation Setup

The evaluations aim to assess the agent's capability to extract structured and accurate information while maintaining a user-engaging interaction. Two testing conditions are designed within the evaluation framework:

- **One-Shot Extraction:** The agent engages with the user to populate the required data fields without posing follow-up questions.
- **Adaptive Extraction:** The agent is allowed to ask clarifying questions to improve the data collection process.

The evaluation presented in this paper was conducted under the assumption that few-shot prompting would enhance the agent's performance by providing it with richer context compared to one-shot prompting [9]. We used a data model $D$, with a broad set of fields, acknowledging that certain fields were unlikely to be completed in a single interaction, particularly for Ambiguous Users. Complex fields, such as nested fields—for example, the history of residential addresses—posed additional challenges. The adaptive setup was expected to mitigate these challenges by allowing iterative clarifications, thereby improving accuracy and reducing ambiguity.

*Ground Truth.* A predefined ground truth profile serves as the reference for evaluation, ensuring alignment between expected and actual outputs. This guides the synthetic user in generating responses, minimizing variability and enabling precise performance assessment.

*Baseline Setup: One-Shot Extraction.* In this setup, the agent attempts to extract information in a single attempt without follow-up queries. Ambiguous or incomplete responses result in data fields being marked as "unclear." This setup serves as a baseline to compare interactive strategies' impact on performance and user engagement. Figure 1 shows an excerpt of the agent's instruction for the baseline condition.

```
### Guidelines for Handling User Responses:
- If a user response lacks a clear, definitive answer or suggests
↪ multiple possibilities, mark the corresponding field as
↪ "unclear" in the data model.
[...]
- Do **not** attempt to resolve ambiguity through follow-up
↪ questions; record the uncertainty and move forward.
[...]
```

Figure 1: Agent instruction for 1-shot extraction

*Adaptive Setup: Follow-Up Questions Allowed.* In this setup, the agent is allowed to ask follow-up questions to clarify unclear responses. It prioritizes user experience by reducing cognitive load, guiding users toward complete and accurate information, maintaining a patient and engaging tone, and minimizing redundant questions. This approach enhances data extraction accuracy and ensures concise, user-focused interactions. Figure 2 shows an excerpt of the agent's instruction for the adaptive condition.

```
1  ### Guidelines for Handling User Responses:
2  - If a user response lacks a clear, definitive answer or suggests
   ↪  multiple possibilities, ask a gentle follow-up question to
   ↪  clarify
3  [...]
4  - You **must** ask follow-up questions until all unclear fields
   ↪  have been clarified.
5  [...]
```

**Figure 2: Agent instruction for adaptive extraction**

## 3.5 Metrics for Evaluation

To compare performance across conditions, we employ key metrics to assess the agent's effectiveness and adaptability: completeness score, correctness score, unclear score, and field-level correctness.

- Completeness Score: Measures data model completion percentage, including fields marked as "unclear." The data collection is complete if all fields are filled, reflecting the agent's information-gathering capability.
- Correctness Score: Evaluates the percentage of correct runs, where the final collected data matches the expected output (as specified in the ground truth), ensuring precision in structured data extraction.
- Unclear Score: Tracks the proportion of fields marked as "unclear," indicating areas where the agent struggles with ambiguous or incomplete user responses.
- Field-Level Correctness: Assesses accuracy at the individual data field level, identifying strengths and weaknesses in data extraction. For instance if the data field *name* is correctly extracted for the 20 runs, the field-level correctness is 100%.

All of these four metrics are measured in percentages from 0 to 100 and as the average across the 20 runs per synthetic user profile and scenario.

## 4 Results

For the purpose of illustrating our framework on a specific use-case, we focus on data collection in the context of a loan application. We evaluated the performance of conversational agents by simulating interactions between the conversational agent and synthetic user profiles. The results of these evaluations are summarized in Figure 3.

### 4.1 Baseline Results

Under the baseline setup, the agent demonstrated robust performance with *Standard Users*, achieving a completion rate of 100% ($\sigma^2 = 0$) and a correctness rate of 80% ($\sigma^2 = 0.17$), as shown in the top-left panel of Figure 3. Most data were accurately captured by the agent, with field-level accuracies reaching 100% across the majority of metrics. However, certain fields, such as those associated with postal codes and years, showed reduced accuracy at 80%. The reason behind the lower extraction performance in postal code fields is due to their nested nature, making one-shot extraction insufficient. The overall unclear percentage was 0%, indicating that the agent effectively handled clear and concise inputs.

In contrast, the agent faced significant challenges when interacting with *Ambiguous Users* in the baseline setup. The correctness rate dropped to 0% ($\sigma^2 = 0.00$), and the overall unclear percentage rose to 55% ($\sigma^2 = 4.42$) (top-left panel of Figure 3), reflecting the difficulty in extracting accurate and complete information from ambiguous inputs. Field-level accuracies varied widely, with high performance in certain straightforward fields, such as email (95%) and full name (100%), but poor results in fields related to loan details, which had 0% accuracy. Notable exceptions included fields like loan term and the boolean debt consolidation field, which achieved 50% and 55% accuracy, respectively. These results highlighted the limitations of one-shot extraction in resolving ambiguity and extracting structured information from inconsistent user responses.

### 4.2 Adaptive Results

The adaptive setup, which allows the agent to ask follow-up questions, significantly improved performance across both user types. For *Standard Users*, the agent achieved a perfect completion rate and correctness rate of 100% ($\sigma^2 = 0.00$), with all data fields populated accurately, as shown in the top-right panel of Figure 3. Both field-level and type-level accuracies reached 100% ($\sigma^2 = 0.00$), confirming the effectiveness of iterative clarification in resolving ambiguities and ensuring accurate information extraction.

For *Ambiguous Users*, the adaptive setup also led to improvements. The correctness rate increased to 60% ($\sigma^2 = 0.27$), and the completion rate rose to 70% ($\sigma^2 = 0.23$), as depicted in the bottom-right panel of Figure 3. Field-level accuracies improved across most data fields, with notable gains observed in fields related to loan details. The overall unclear percentage for Ambiguous Users was reduced to 0%, demonstrating the agent's ability to clarify and resolve ambiguities effectively through adaptive interactions.

Despite these improvements, performance gaps persisted between both user types in the adaptive setup. *Standard Users* consistently outperformed *Ambiguous Users*, with correctness and completion rates higher by 40% and 30%, respectively (bottom-left and bottom-right panels of Figure 3). Certain data fields, particularly those requiring nuanced interpretation, such as residence history, continued to present challenges for the agent when interacting with *Ambiguous Users*. One could significantly improve the performance by re-applying the evaluation to alternative prompts in the "Adaptive set up" (e.g., increased user guidance).

## 5 Limitations and Future Work

This study presents a framework for evaluating conversational agents, with a focus on structured, task-oriented interactions. While it shows promising results, limitations and directions for future work should be considered. Firstly, it is important to note that the loan application scenario is used as an illustrative example to introduce and demonstrate the framework. While this example provides a structured environment with well-defined data fields and task objectives, it does not represent the full spectrum of potential applications for conversational agents. Applying the framework to other domains (e.g., legal, healthtech, education) presents additional challenges that are not addressed here. Another limitation is the simplification of the example proposed in this study. For the purpose of demonstrating the loan application scenario, we used only a small fraction of the data that would typically be required in a real-world application. However, due to the scalable and flexible nature of the framework, one can easily adapt it across different applications and domains. Expanding the framework to handle larger datasets and more complex interaction scenarios will further validate its robustness and applicability in diverse contexts. These limitations highlight the need for caution when interpreting results beyond the specific contexts tested in this study. Additionally, the current evaluation primarily focuses on accuracy in information extraction and task completion, with limited focus on conversational flow and user engagement. Future work will aim to incorporate additional evaluation metrics, such as Turn-Level Appropriateness, Dialogue Coherence, and Reengagement Rate, to better assess and improve the agent's conversational effectiveness. The adoption of automated evaluation methods, including transformer-based models (e.g., BERT, RoBERTa) and entailment-based approaches, presents an opportunity to achieve scalable and precise insights into these aspects [11]. By building on this work and extending the framework to a wider range of applications and evaluation metrics, our ongoing research will aim to refine the evaluation of conversational agents. Incorporating additional metrics for conversational flow and user engagement will ensure that these systems are not only capable of meeting structured information extraction goals but also of providing intuitive and engaging conversational experiences by ensuring better user engagement and experience. Our primary objective is to share early-stage contributions and ideas to foster discussion among the community while laying the groundwork for our future efforts to further develop and expand this work.

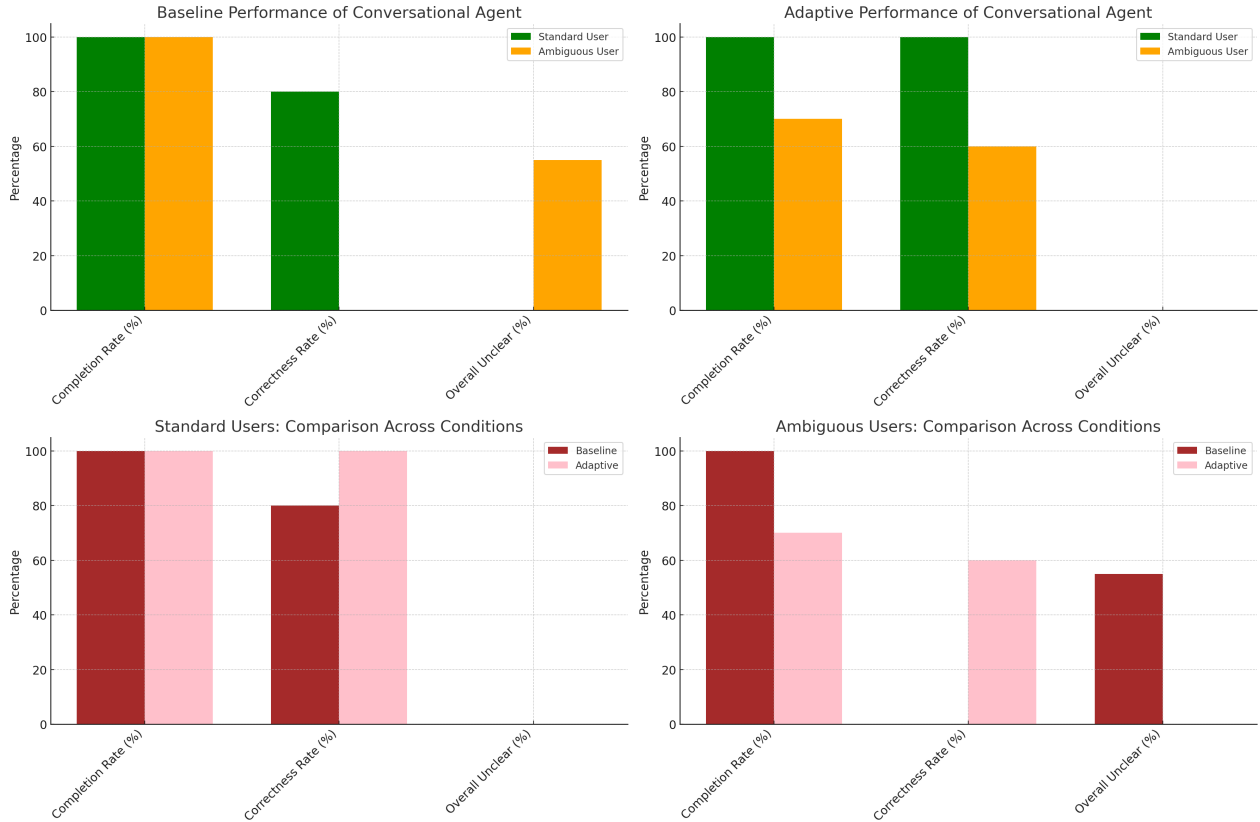

Figure 3: Summary of evaluation results

## 6 Conclusion

In this paper, we presented an initial exploration of a dynamic benchmarking framework aimed at evaluating large language model (LLM)-based conversational agents through interactions with synthetic users. Our approach aims to address some of the challenges associated with evaluation methods, which often fail to capture the nuances of user experience [10]. Furthermore, the ability to interact with LLMs via different modalities complicates evaluation, necessitating new benchmarks to assess cross-modal integration capabilities [25]. Our experimental results demonstrate that while conversational agents perform well at extracting data from clear and straightforward conversations, their performance significantly improves when adaptive strategies are employed to handle ambiguous users. This highlights the importance of iterative questioning in improving information accuracy and completeness in conversational data capture. Our evaluations show that the adaptive approach improves correctness and reduces unclear data, especially with ambiguous users. Building on these findings, future work will focus on extending the framework to broader domains, incorporating additional user types, and exploring advanced evaluation metrics such as dialogue coherence and user engagement. In conclusion, our dynamic benchmarking framework offers a flexible, multi-faceted, and scalable approach for evaluating LLM-based conversational agents in simulated real-world contexts.

## A Pseudo code

**Algorithm 1** Simulation Framework for Conversational Agents

**Require:** Data $D = \{f_1, f_2, \ldots, f_n\}$, Max steps $T$, Agent prompt function, Synthetic user function
**Ensure:** Populated data $D$ or termination reason
1: **Initialization:**
2: Initialize the agent with a prompt based on the empty data: $P_0 = \text{generate\_prompt}(D)$
3: Initialize the user with an initial synthetic message: $u_0 = $ "Hello, I am ready."
4: Save initial user message: $H = \{u_0\}$
5: **Main Loop:**
6: **while** NOT **STOP do**
7:   **if** NOT first interaction **then**
8:     Ask the agent for a message in response to the user: $a_t = \text{generate\_response}(P, u_t)$
9:     Save the user's message: $H = H \cup \{u_t\}$
10:   **end if**
11:   Agent generates a reply based on the current data and user message:
12:         $a_t = \text{generate\_response}(P, H)$
13:   Save the agent's reply: $H = H \cup \{a_t\}$
14:   Parse the agent's reply to update the data: $D = \text{update\_data}(D, a_t)$
15:   **Check Stopping Conditions:**
16:   **if** Data $D$ is fully populated **OR** maximum steps $T$ exceeded **then**
17:     Set **STOP** to true
18:   **end if**
19: **end while**
20: **Termination:**
21: Return the data $S$ or termination reason.

## B User Profile

- **Email:** A string representing the applicant's email address.
- **Annual Income:**
  - *Amount:* A numerical value representing the income amount.
  - *Currency:* A string indicating the currency of the income.
- **Last Name:** A string containing the applicant's last name.
- **First Name:** A string containing the applicant's first name.
- **Postal Code History:**
  - A list of past and current postal codes, each with:
    * *Year:* A numerical value representing the year.
    * *Postal Code:* A string representing the postal code.
  - Up to the last three postcodes of residency.
- **Date of Birth:** A string formatted as DD-MM-YYYY.
- **Phone Number:** A numerical value representing the applicant's phone number.
- **Employment Status:** A string indicating one of the following options:
  - Employed
  - Self-employed
  - Retired
  - Student
- **Residential Status:** A string indicating one of the following options:
  - Renting
  - Living with parents

## C  Loan Details

- **Loan Amount:**
  - *Amount:* A numerical value representing the loan amount.
  - *Currency:* A string indicating the currency of the loan.
- **Loan Purpose:** A string representing the purpose of the loan, which can be one of the following:
  - Home renovation
  - Debt consolidation
  - Education
  - Medical expenses
  - Vacation
- **Loan Term (Months):** A numerical value representing the loan duration in months.
- **Debt Consolidation Indicator:** A boolean value indicating whether the loan is for debt consolidation.